\pdfoutput=1

\documentclass[11pt]{article}

\usepackage[final]{acl_new}

\usepackage{times}
\usepackage{latexsym}
\usepackage{enumitem}
\usepackage{comment}
\usepackage{xspace}

\usepackage[T1]{fontenc}

\usepackage[utf8]{inputenc}

\usepackage{microtype}

\usepackage{inconsolata}

\usepackage{times}
\usepackage{latexsym}
\usepackage{booktabs}
\usepackage{mdwlist}

\usepackage{inconsolata}
\usepackage{caption}
\usepackage{subcaption}
\usepackage{multirow}
\usepackage{graphicx}
\usepackage{tabularx} 
\usepackage{wrapfig}
\usepackage{amsmath} 
\usepackage{xcolor} 
\usepackage{tcolorbox} 
\usepackage{algorithm}
\usepackage{algpseudocode}

\newcommand{\envname}{\textsc{CoBlock}\xspace}

\newcommand{\prompttext}[1]{{\textit{#1}}\\}

\definecolor{boxbackground}{RGB}{225, 225, 255} 

\newtcolorbox{reasoningbox}[1][]{
  colback=boxbackground,
  colframe=black,
  boxrule=0.5pt,
  arc=4pt,
  boxsep=5pt,
  left=5pt,
  right=5pt,
  top=5pt,
  bottom=5pt,
  #1
}
\newcommand{\createpromptbox}[2]{
  \begin{reasoningbox}
    \textbf{#1}\\
    #2
  \end{reasoningbox}
}

\newcommand{\inpromptinventory}[2]{
    \textit{<block block$\_$type =``#1'', count=#2>}
}

\newcommand{\inpromptblock}[2]{
    \textit{<block block$\_$type =``#1'', pos=#2>}
}

\title{Your Co-Workers Matter: Evaluating Collaborative Capabilities of Language Models in Blocks World}

\author{Guande Wu\textsuperscript{1}, Chen Zhao\textsuperscript{1,2}, Claudio Silva\textsuperscript{1}, He He\textsuperscript{1} \\
  \textsuperscript{1}New York University, \textsuperscript{2}NYU Shanghai \\
  {\{guandewu, cz1285, csilva, hhe\}@nyu.edu} \\}

\begin{document}
\maketitle
\begin{abstract}


Language agents that interact with the world on their own have great potential for automating digital tasks.
While large language model (LLM) agents have made progress in understanding and executing tasks such as textual games and webpage control, 
many real-world tasks also require collaboration with humans or other LLMs in equal roles, which involves intent understanding, task coordination, and communication.
To test LLM's ability to collaborate, we design a blocks-world environment,
where two agents, each having unique goals and skills, build a target structure together. To complete the goals, 
they can act in the world and communicate in natural language. 
Under this environment, we design increasingly challenging settings to evaluate different collaboration perspectives, from independent to more complex, dependent tasks.
We further adopt chain-of-thought prompts that include intermediate reasoning steps to model the partner's state and identify and correct execution errors. 
Both human-machine and machine-machine experiments show that
LLM agents have strong grounding capacities, and our 
approach significantly improves the evaluation metric.

\end{abstract}

\section{Introduction}
\begin{figure}[!ht]
\centering
\includegraphics[width=\linewidth]{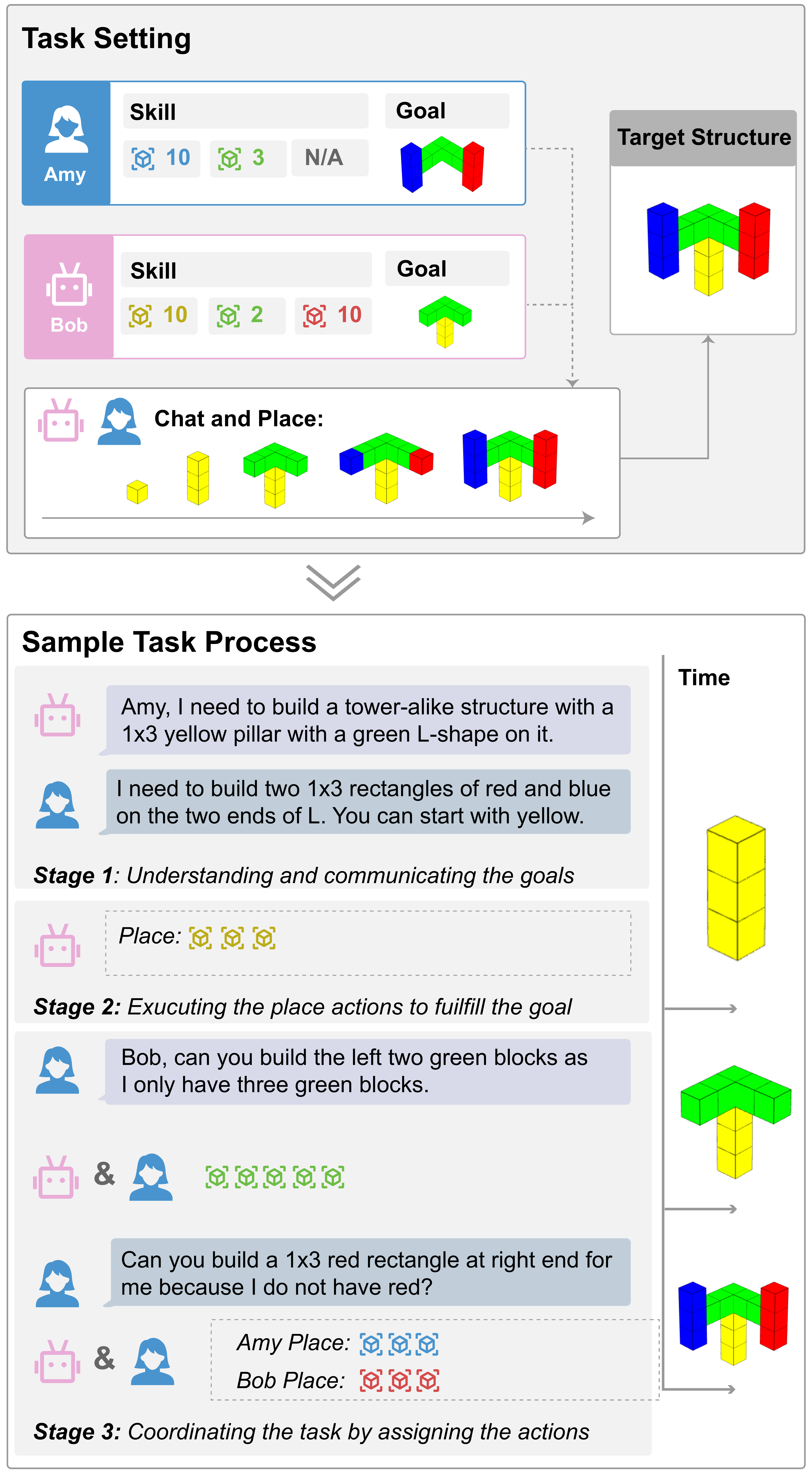} 
\caption{\textbf{Task Setting: } A human agent (Amy) and an LLM agent (Bob) collaborate on building the block structure with diverse goals and inventories. \textbf{Sample Task Process: } In the shown task, Amy's goal relies on Bob's, so they have to coordinate. To succeed on this task, Amy and Bob have to 1) communicate their goals and figure out the immediate plan to complete; 2) Amy \textit{place} the yellow blocks to complete the immediate plan; 3) Amy and Bob coordinate to complete the remaining part of their goals. 
}
\label{fig:fig1}
\end{figure}
As large language models (LLMs) evolve, they are increasingly expected to collaborate closely with humans or other LLM agents, emphasizing the importance of coordination and communication~\cite{radford2019language}.
For instance, an LLM assistant might need to work alongside human professionals to plan tasks, manage projects, and ensure efficient execution.
While multi-agent collaboration is not a new area~\cite{leibo2021meltingpot,carroll2019utility,he2017symmetric}, there is limited study on how LLM agents collaborate with humans in {\it equal} roles, rather than passively following human's instructions.
To fill this gap, we introduce a collaborative blocks world environment (\envname), to evaluate the collaboration ability of LLMs. 
In this environment, two agents (either human or LLM), each equipped with complementary goals and skills, collaborate to build a target structure without a designated leader. For example, as shown in Figure~\ref{fig:fig1}, Amy is responsible for the top structure, which must be built on top of the base structure that Bob needs to build first. Meanwhile, Amy does not have the red blocks that are necessary to complete her goal. Therefore, they must coordinate their actions, like deciding what to build first and who builds what (e.g., Amy may ask Bob to build the yellow pillar first).

\begin{figure*}[ht]
\centering
\includegraphics[width=\linewidth]{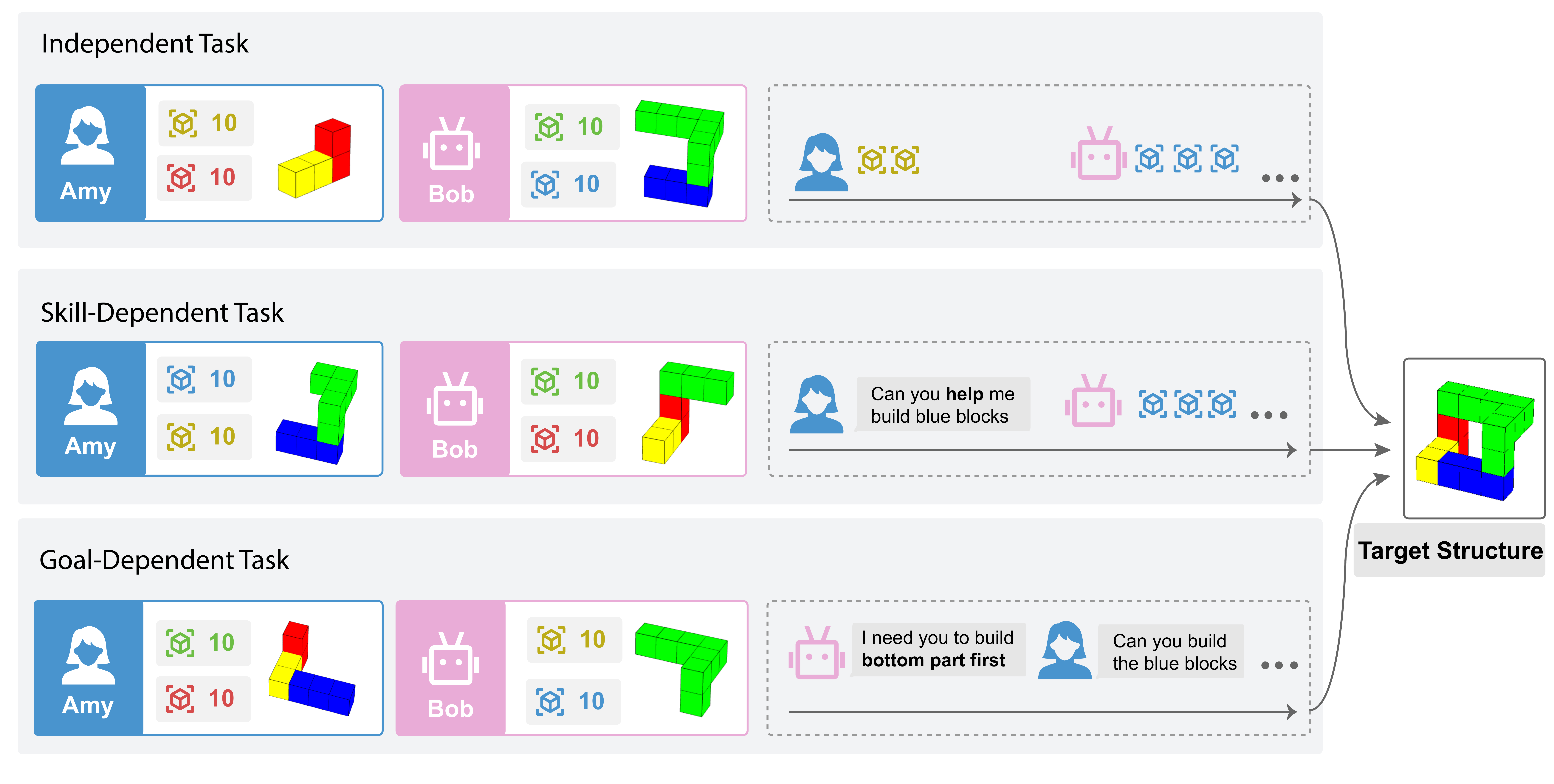} 
\caption{Three different collaboration tasks with increasing levels of coordination. Top: The independent tasks that require little coordination between agents; Middle: The skill-dependent tasks that at least one goal requires both agents to complete; Bottom: The goal-dependent tasks that one agent's goal depends on prior completion of the partner’s goal.}

\label{fig:task}
\end{figure*}

We design three scenarios to progressively increase the level of interaction and coordination required between the two agents, moving from independent tasks to more complex, interdependent tasks (Figure~\ref{fig:task}).
\begin{itemize}[itemsep=0pt]
    \item {\bf Independent tasks}: Each agent can complete its assigned partial structure without the partner's participation. 
    \item {\bf Skill-dependent tasks}: One agent relies on the other's ``skill'' or resources to complete the task. For instance, Bob needs Amy's help to build the yellow bar because he doesn't have any yellow blocks. 
    \item {\bf Goal-dependent tasks}: This scenario involves dependence between the agents' assigned partial structure (i.e. goal), requiring more advanced planning and coordination. 
\end{itemize}
To ground LLMs in~\envname and enable them to complete the tasks, we prompt GPT series models (GPT-3.5 and GPT-4) to make decisions about the next action given the current observation. 
At each round, our basic prompt describes the agent's goal (the color and position of every block), the current state (the structure built so far, the action and message history), and candidate actions to choose from (place a block, send a message, wait or terminate the task).

To guide the model in choosing the right action, we additionally include chain-of-thought prompts (CoT)~\cite{DBLP:conf/nips/Wei0SBIXCLZ22} to go through intermediate reasoning steps:
(1) analysis of the currently built structure and the agent's goal;
(2) partner-state modeling to predict the current intent and state of the partner;
and (3) self-reflection that identifies and corrects any execution error, and adjusts communication style accordingly. 

We evaluate the LLM agents in~\envname~under both machine-machine and human-machine settings, measured by task success rate,
total steps,
and workload balance. 
First, we find that LLM agents have a high success rate when building the structure independently,
demonstrating strong grounding ability even without multi-modal training.
Second, adding partner-state modeling and self-reflection prompting results in a 30\% absolute increase in task completion rate and leads to better workload balance.
Finally, while human-machine collaboration has a slightly higher success rate than machine-machine collaboration, humans often take on more responsibility in challenging scenarios when the LLM agent struggles.
We hope our findings and the evaluation environment will support future studies on communication and coordination in multi-agent collaboration.\footnote{The code and data is available at https://github.com/jnzs1836/coblocks.}

\section{The Collaborative Blocks World}
In this section, we describe the collaborative blocks world environment~\envname{} (Section \ref{subsec:setting}) and the three types of collaboration tasks (Section \ref{subsec:task}).

\label{sec:task}

\subsection{The \envname{} Environment}
\label{subsec:setting}

As shown in Figure~\ref{fig:fig1}, in each task, two agents work together to build a target structure by taking actions in the \envname{}. Specifically, either agent takes one action in each round, we move to the next round after actions from both agents are executed. The task is terminated when the target structure is completed or either agent takes the \textit{end\_task} action.

\paragraph{Structure.}
A {\it structure} $G$ consists of a set of blocks $B=\{b_1, b_2,\cdots, b_m\}$, where each block $b$ is specified by a color $c$ from a set of colors $C$ (we use a total of six colors), 
and a 3D position given by coordinates $(x, y, z)$. 
To simulate real-world conditions, a structure must adhere to the gravity restriction, where each block must be adjacent to at least one other block or be on the ground plane (i.e.\ $y=0$).
A {\it sub-structure} of $G$ is a smaller structure formed from a subset of $
G$'s blocks.

\paragraph{Agent.}
Each agent is given a {\it goal} $g$, which is a sub-structure of the target structure $G$. 
The union of the sub-structures forms the target structure.
For example, in Figure~\ref{fig:fig1}, Amy is given the upper part of the target structure while Bob is given the lower pillar.
Additionally, each agent has an {\it inventory} $e = \{(c_1, n_1), \cdots, (c_k, n_k)\}$, where $n_i$ denotes the number of blocks of color $c_i$. Each inventory contains a subset of colors present in $G$. 
The goal and inventory of each agent are private so they must communicate to figure out who needs what.

\paragraph{State.}
The state of the world consists of all public information, including the structure built so far, and the history of the agents' utterances and actions. 

\paragraph{Actions.}
An agent can take one of the following actions: {\it place} a block from its inventory at specific coordinates; {\it remove} a block that was placed at specific coordinates (the block will not be used anymore);
{\it send messages} to its partner; {\it wait} without performing any action;
and {\it terminate} the task.

\subsection{Collaboration Tasks}
\label{subsec:task}

To evaluate different aspects of collaboration, we design three types of tasks with increasing levels of dependence between two agents. We ensure there exists at least one solution to complete the task.
Examples of the tasks are shown in Figure~\ref{fig:task}.



\paragraph{Independent tasks.}
These tasks require minimal coordination, as
two agents can build their goals separately to complete the target structure.
However, the agents still need to communicate initially to determine that they can each build their substructures independently.

\paragraph{Skill-dependent tasks.}
An agent cannot complete its goal without assistance from its partner, as it lacks certain colors of blocks.
%
For example, in Figure~\ref{fig:task}, 
Amy needs to describe the green sub-structure to Bob so that Bob can build it using his green blocks. To succeed in these tasks, an agent must communicate its needs, understand the partner's needs, and then coordinate the actions accordingly.


\paragraph{Goal-dependent tasks.} 
In these tasks, in addition to the dependence between agents' skills,
an agent's ability to complete their goal may depend on prior completion of the partner's goal due to the gravity restriction.
For example, in Figure~\ref{fig:task}, Amy must build her sub-structure first (which requires Bob's help with the yellow blocks), so that Bob can build his sub-structure on top of it later.
These tasks are the most challenging, as there are complex dependencies among the steps to finish the target structure.

\section{Building Collaborative Agents based on LLMs}

\label{sec:method}

\begin{figure*}[ht]
\centering
\includegraphics[width=\linewidth]{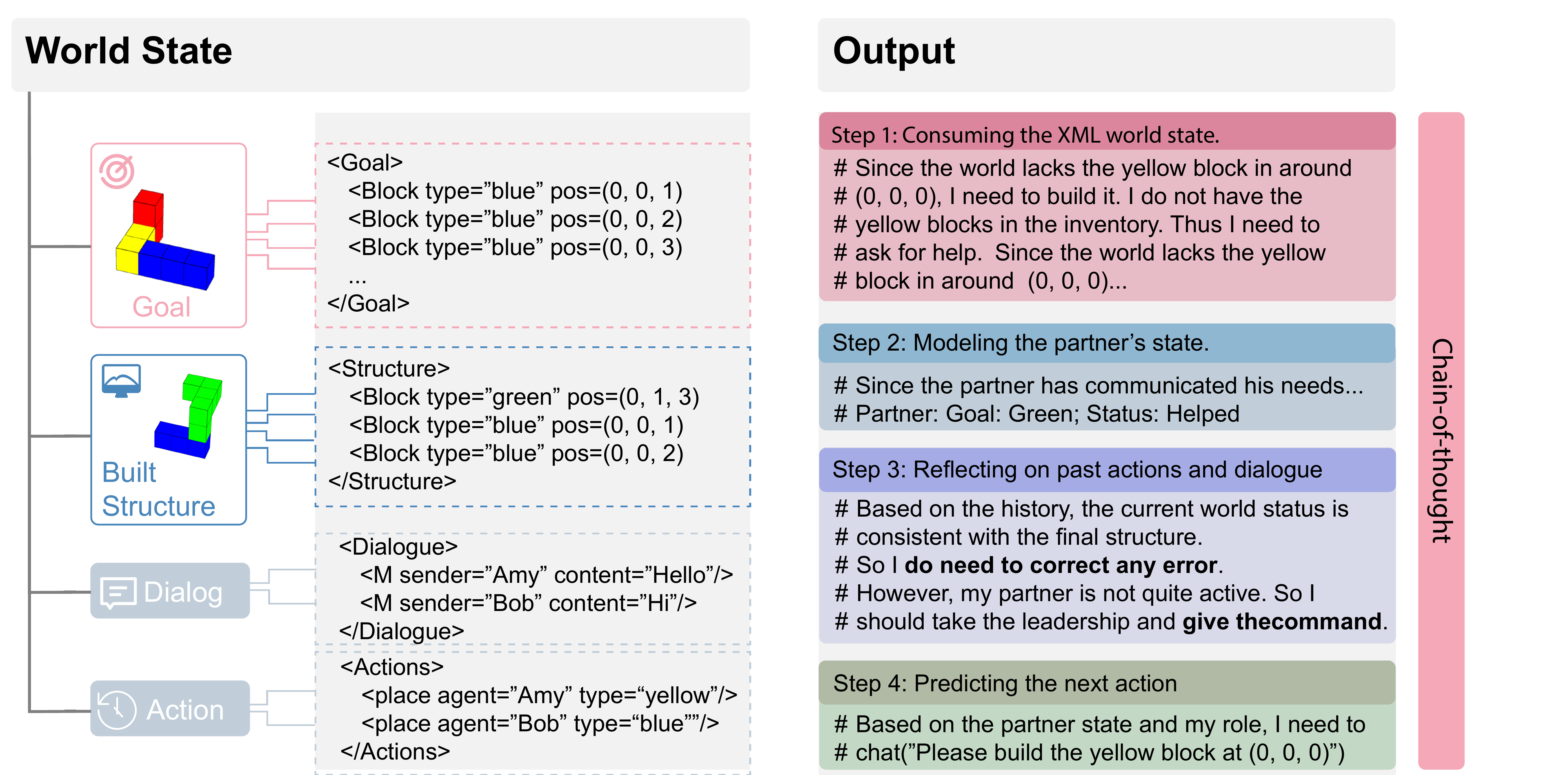} 
\caption{\textbf{World State} consists of the agent's goal, currently built structure, dialogue, and action history. \textbf{Prompt Text} consists of four steps: 1) Analyze the XML world state and summarize the useful information; 2) Infer both the agent and the partner's state; 3) Self-reflection which identifies the errors and adjusts the communication strategies; 4) Predict the action. We use the CoT prompts in all steps.} 
\label{fig:cot}
\end{figure*}

We use an LLM as the base agent and query it at each round to predict the next action given the current world state (e.g., the structure built so far and the agent's goal). 
\begin{itemize}[itemsep=0pt]
    \item {\bf Input:} The input consists of descriptions of the state of the world described in XML (Figure~\ref{fig:cot} left),
including the goal of each agent, the structure built so far, the dialogue and action histories.
We use a sequence of blocks (with color and 3D position) to represent the structures.
\item {\bf Output:} The output is the next action (\textit{place,
break, send message, wait or terminate}).
We instruct the model to use a specific format
(e.g., {\tt send\_message(message="Hello")})
to ensure that it can be parsed for execution in the environment.
\end{itemize}


To help the LLM agent decide the next action, we use CoT prompting to guide the LLM agent through several intermediate steps designed to anticipate the partner's needs and adjust the construction plan given new observations. 
We describe each step below and show examples in Figure~\ref{fig:cot}.

%




\paragraph{Step 1: understanding the world state.}
The world state is represented using the XML format.
Therefore, we prompt LLM agent to first parse the state and describe it in natural language.
For example, given the {\tt <goal>} and {\tt <built structure>} field, the agent infers that it should build the yellow blocks next.

\paragraph{Step 2: modeling the partner's state.}
Effective collaboration requires an agent to understand the partner's needs and coordinate with them to strike a balance between assisting others and fulfilling their own goal.
For example, in skill- and goal-dependent tasks, the agent must identify which sub-structures their partner needs help with and integrate it into the planning of their goals.
Therefore, we incorporate 
{\it partner-state modeling}~\cite{DBLP:conf/uai/Boutilier96, DBLP:conf/atal/ChalkiadakisB03} into our prompts as an intermediate step.
%
As shown in Figure~\ref{fig:cot}, based on the dialogue and action history, the agent predicts the intent and state of its partner, including the goal, the inventory set, the immediate next plan (e.g., building the green sub-structure), and whether the plan has been executed.

\paragraph{Step 3: reflecting on past actions and dialogue.}
Misunderstandings in collaboration can lead to incorrect actions, and if these errors aren't corrected promptly, they may result in higher expenses to rectify them later. For instance, an agent might place a block of the wrong color at a specified position. If this error isn't corrected in the subsequent round, it could take multiple backtracking actions to correct the mistake.
To address this issue, we adopt the {\it self-reflection mechanism~}\cite{DBLP:journals/corr/abs-2305-10142} to help the agent review the currently built structure and correct low-level construction errors and high-level communication strategies. 

Specifically, we compare the built structure and the agent's goal to check if there is an error using an external program (i.e.\ whether the built structure is exactly a partial structure of the goal).
The feedback is then sent to the agent to correct the potential error. 
In addition, we instruct the agent to adjust their communication strategies~\cite{hovland1953communication} based on the feedback along  three dimensions:
team role (whether to take the lead and assign tasks to the partner),
altruism/egoism balance (whether to pursue their own goals or assist the partner), and persuasion strategy (whether to be more proactive and persuasive in communication).



\paragraph{Step 4: predicting the next action.}
The final step generates the next action based on the world state, partner state, and the reflection result in previous steps.

\paragraph{}
Our final prompt starts with a general description of the game and the input/output format for the model to understand the XML and the action functions.
To guide the LLM agent to decide the next action through the above reasoning steps,
we then provide a CoT prompt that consists of a sequence of reasoning examples,
where each example consists of the current world state and the four reasoning steps.
A full prompt is shown in Appendix~\ref{sec:prompt}.

\section{Experiments}


Successfully completing tasks in \envname requires both grounding and collaboration abilities.
Therefore, we first access whether LLMs are able to execute instructions in \envname correctly through a \textbf{single-agent experiment}  (Section~\ref{subsec:single-exp-setting}).
Next, we evaluate whether LLM agents can effectively collaborate with other LLM  or human agents through communication and coordination using a \textbf{multi-agent experiment} (Section~\ref{subsec:multi-exp-setting} and \ref{subsec:multi-exp-result}).

\subsection{Implementation Details}

\paragraph{Environment.}
We implement the~\envname environment using a web interface based on React.js~\cite{React}, and render the 3D world through THREE.js~\cite{danchilla2012three}. 
To generate target structures, we follow five common structures used in~\cite{DBLP:conf/acl/Narayan-ChenJH19}: symbol, bridge, arch, tower, and rectangle.
We then use depth-first search to find combination of these common structures and manually adjusted them if needed.
To generate the goals and skills for each agent, we manually split the target structures into two sub-structures and assign inventories to make sure that the tasks can be completed collaboratively.
In total, we create 24 collaboration tasks in \envname.
The details of task creation can be found in Appendix~\ref{supp:structure}. 

\paragraph{Agents.}
We recruit human participants with experience playing Minecraft (which is similar to \envname) via Amazon Mechanical Turk. Participants must complete a tutorial and pass a qualification exam to perform the tasks.
We pay participants with an hourly rate of \$18.
Based on our pilot study, the tasks are usually completed within thirty minutes.
For LLM agents, we provide eight in-context CoT examples covering three types of tasks: two  independent tasks, three  skill-dependent tasks, and three goal-dependent tasks.
we also include a baseline LLM agent, using
CoT prompts but without partner-state modeling (Step 2) and self-reflection mechanism (Step 3). 


\subsection{Single-Agent Experiment}
\label{subsec:single-exp-setting}


\begin{figure*}[ht]

\begin{center}
    
\includegraphics[width=0.99\textwidth]{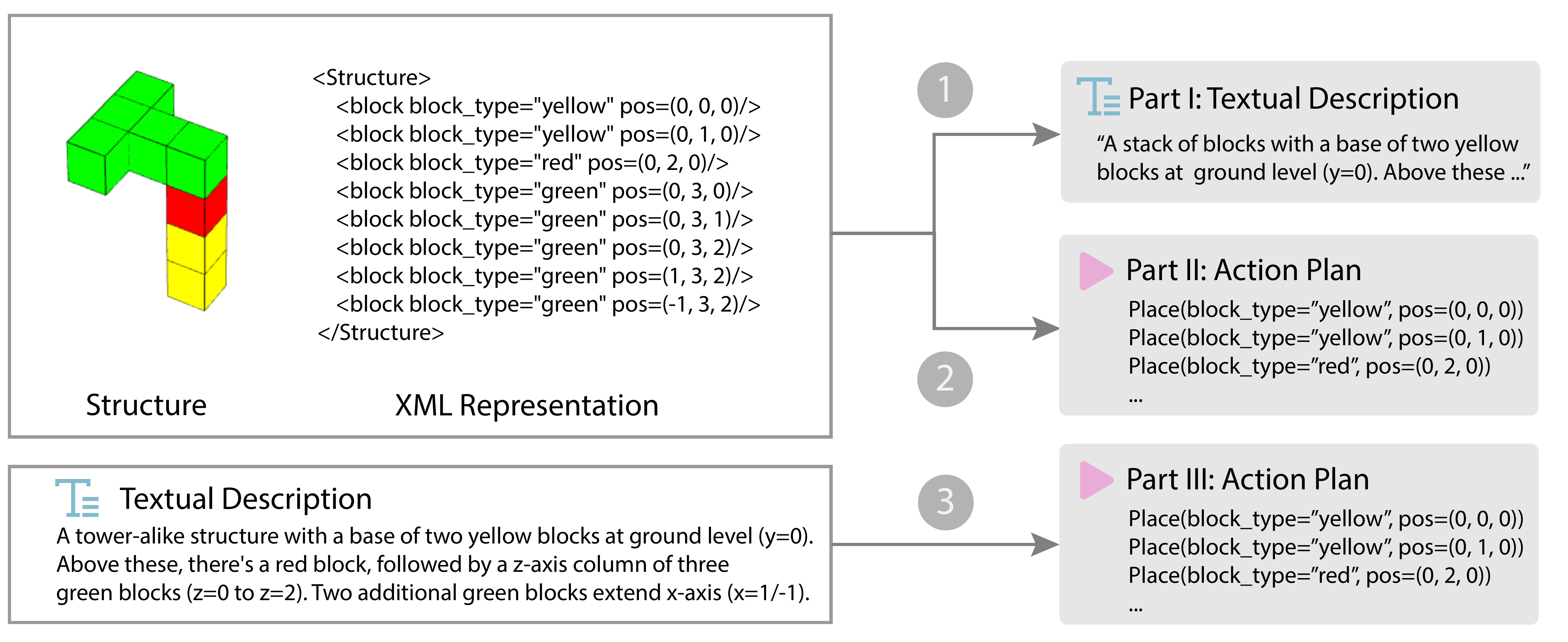} 
\caption{Single-agent experiment settings including three parts. We represent the blocks by the XML structure and the textual description. \textbf{Part I}: describe the given XML into textual descriptions. \textbf{Part II}: convert the XML into a sequence of commands. \textbf{Part III}: directly convert the textual description into a sequence of commands.}
\label{fig:preliminary}
\end{center}

\end{figure*}
We propose three single-agent tasks to evaluate the grounding abilities of LLM agents (Figure~\ref{fig:preliminary}): 1) Generate textual descriptions from XML-represented structures; We manually 
evaluate these descriptions to verify the LLM agent's ability to describe the target structure. 2) Generate a sequence of actions given the XML structures; This task measures the agent's ability for task planning. 3) Generate the action sequence from the textual descriptions. This task combines the first two tasks and therefore requires both grounding and planning capacities.

\noindent \textbf{Evaluation Metrics.} We use \textbf{success rate} to evaluate the proportion of tasks completed by agents successfully (i.e. generate the correct description in task 1 and the action plan in tasks 2 and 3):
\begin{equation}
    \text{Success Rate} = \frac{\text{Number of Success
}}{
\text{Total Number of Tasks}
}
\label{eq:success-rate}
\end{equation}

\begin{figure}[ht]

\begin{center}
    
\includegraphics[width=0.45\textwidth]{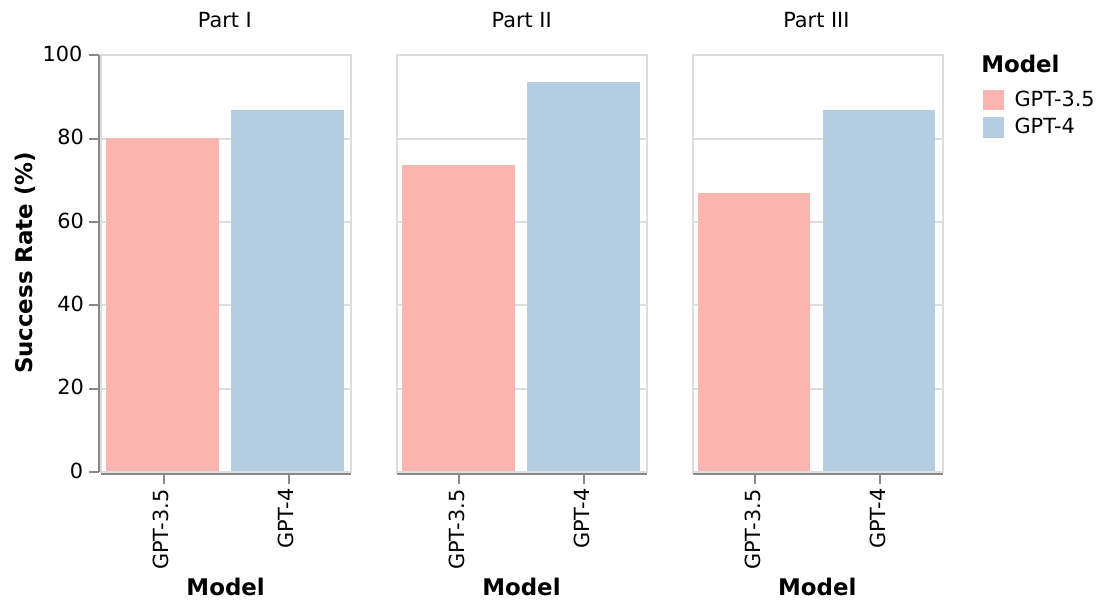} 
\caption{Experiment results on single-agent experiments (Part I, II, III). LLM agents successfully complete almost all tasks.}
\label{fig:grounding}
\end{center}

\end{figure}

\noindent \textbf{Results.} 
According to Figure~\ref{fig:grounding}, both GPT-4 and GPT-3.5 agents successfully complete almost all three tasks, which 
indicates that LLM agents have the sufficient grounding and planning skills for the collaborative blocks world. Therefore, 
the observed failures in the collaboration tasks (which we present next) should not mainly come from the grounding and planning capacities of agents.

\subsection{Multi-Agent Experiment Setup} 
\label{subsec:multi-exp-setting}

We propose two multi-agent task settings: \textbf{Human-machine collaboration} that a human agent and an LLM agent build the structures. This setting aims to evaluate the ability of LLM agents to act as proactive collaborators that fulfill their own goals while \textit{assisting} human partners;
\textbf{Machine-machine collaboration} that two LLM agents build the structures, primarily evaluates the collaboration behaviors between two LLM agents.
\noindent \textbf{Evaluation Metrics.}
Similar to the single-agent setting, we use success rate (See Equation~\ref{eq:success-rate}) to measure whether agent(s) can complete the target structure. 
We propose two additional metrics targeting the effectiveness of collaboration~\cite{mcewan2017effectiveness} during task execution:
%
%
%

\noindent \textbf{Workload Balance} measures how tasks are distributed evenly between two agents. 
Since the agents have varying goals and skills, even optimal task assignments result in different action numbers. Therefore, we normalize the number of actions by comparing them with the optimal action number 
\begin{equation}
    a_1 = \frac{N_a \cdot N^*_b}{N^*_a}; b = N_b
\end{equation}
where the $N_a$ and $N_b$ refer to the numbers of actions and $N^*_a$ and $N^*_b$ are the numbers of optimal actions. The final workload balance $\gamma$ is computed as
\begin{equation}
    \gamma = \frac{a * b}{a^2 + b^2}.
\end{equation}
The optimal value of $\gamma$ is $0.5$.

\noindent \textbf{Task Completion Timesteps} is defined as the total number of actions required to complete the task. \footnote{We do not use wallclock time because it can be affected by the network latency and the participant's response time.}

\begin{figure*}[ht]

\begin{center}
    
\includegraphics[width=\textwidth]{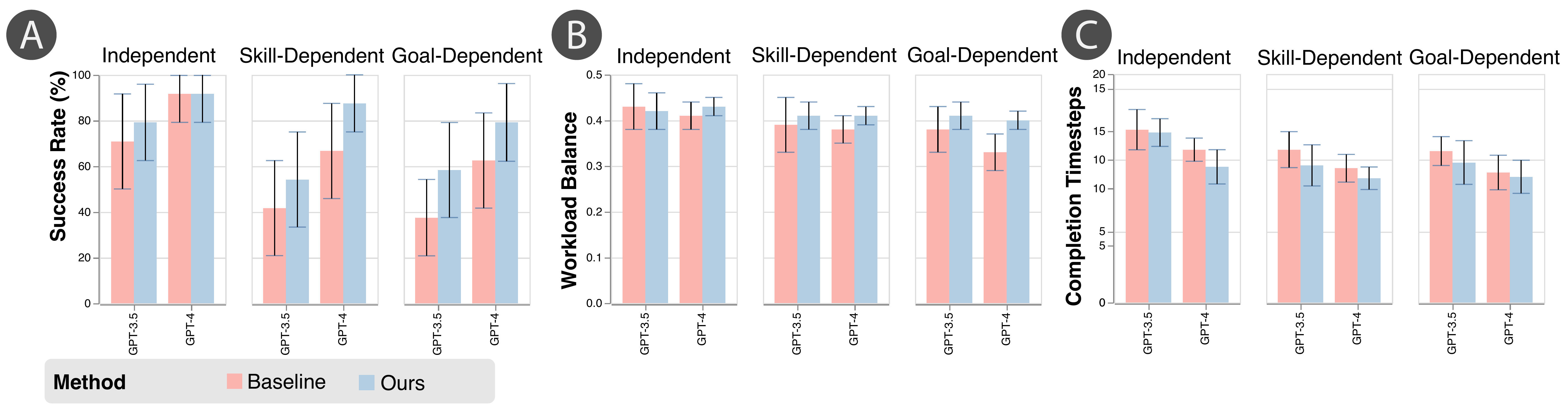} 
\caption{Human-machine experiment results. The experiments are conducted on the independent, skill-dependent, and goal-dependent tasks with both GPT-3.5 and GPT-4. We include both LLM agents and baseline LLM agents without partner-state modeling and reflection.}
\label{fig:human-model-all}
\end{center}

\end{figure*}

\begin{figure*}[ht]

\begin{center}
    
\includegraphics[width=\textwidth]{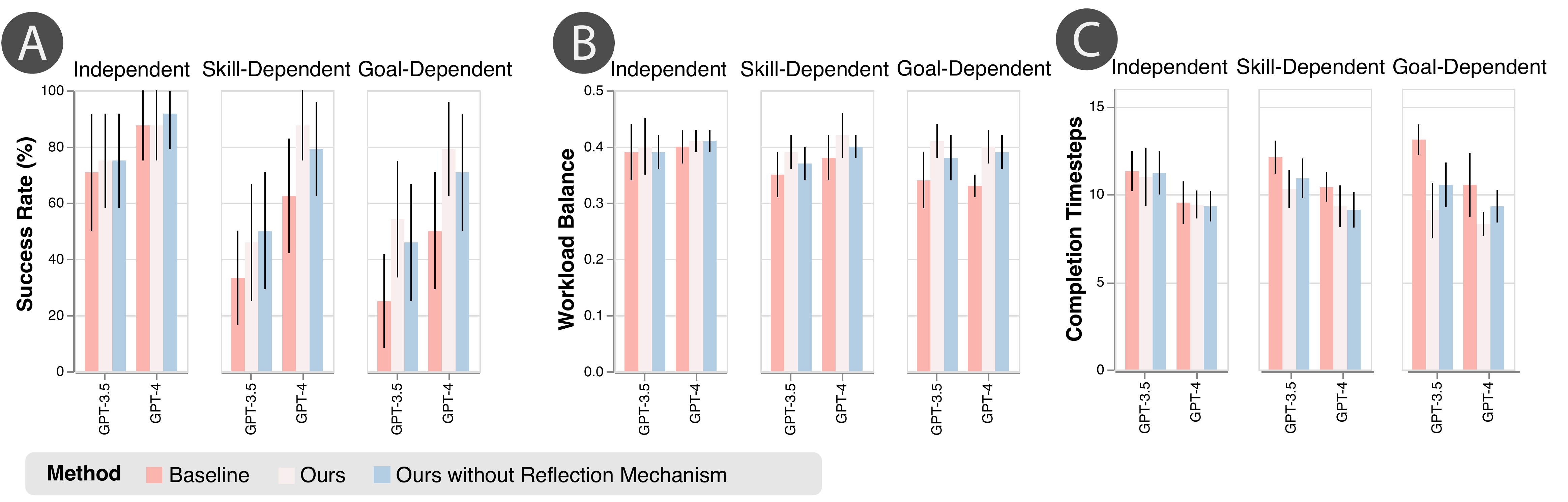} 
\caption{Machine-machine experiment results. Similar to human-machine experiments, we include independent, skill-dependent, and goal-dependent tasks with both GPT-3.5 and GPT-4. We include LLM agents, LLM agents without reflection and baseline LLM agents without both partner-state modeling and reflection.
    }
\label{fig:model-only-all}
\end{center}

\end{figure*}

\begin{figure}[ht]

\begin{center}
    
\includegraphics[width=0.40\textwidth]{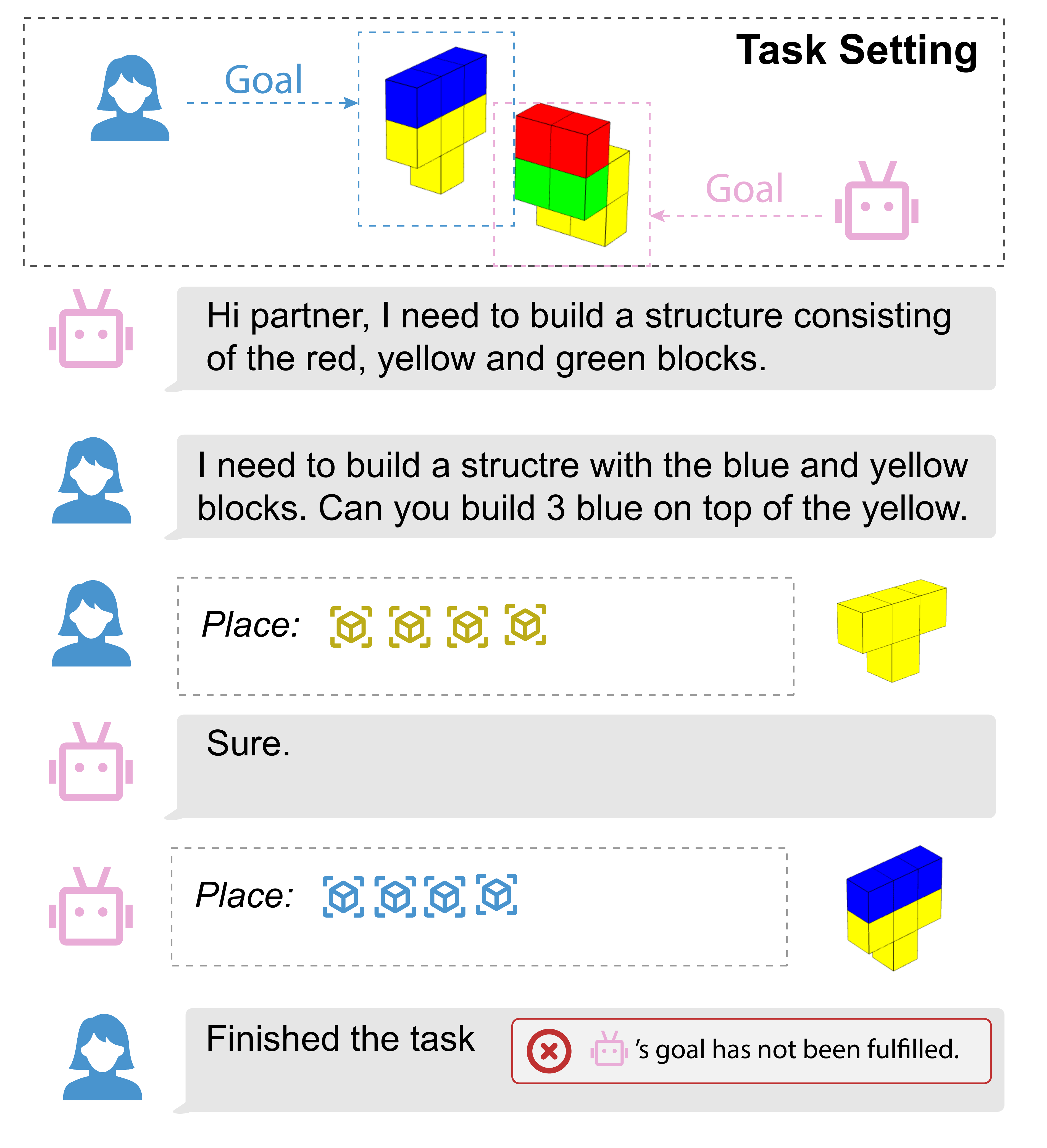} 
\caption{A human-machine example in the skill-dependent task.  Our baseline LLM agents fail to complete their own goals due to the priority of assisting the partner. Intermediate steps are omitted for clarity.}
\label{fig:exp-example-1}
\end{center}

\end{figure}
\subsection{Multi-Agent Experiment Results}
\label{subsec:multi-exp-result}

We present the human-machine experiment results in Figure~\ref{fig:human-model-all} and machine-machine results in Figure~\ref{fig:model-only-all}. 
In addition to quantitative results, we conduct an error analysis by manually collecting all failure instances (73 in total) 
from both human-machine and machine-machine experiments. 



\noindent \textbf{Baseline LLM agents struggle to complete the task. } 
According to Figure~\ref{fig:human-model-all} (A)  and Figure~\ref{fig:model-only-all} (A), the baseline agents suffer from a low task success rate in the skill-dependent and goal-dependent tasks. The contrast with single-agent experiment results suggests that collaboration presents significant challenges. We find that LLM agents often prioritize assisting their partners and neglecting their own goals (55.6\% errors in the goal-dependent tasks and 37.5\% errors in the skill-dependent tasks).
In addition, LLM agents lack initiative in seeking additional information from the partner (11.1\% errors in the skill-dependent tasks and 25.0\% errors in the goal-dependent tasks).
For instance, in Figure~\ref{fig:exp-example-1}, when a human's intention (build blue blocks) conflicts with the LLM agent (build green blocks), the LLM agent stops executing its own goal, but decides to help human instead.



\begin{figure}[!ht]

\begin{center}
    
\includegraphics[width=0.5\textwidth]{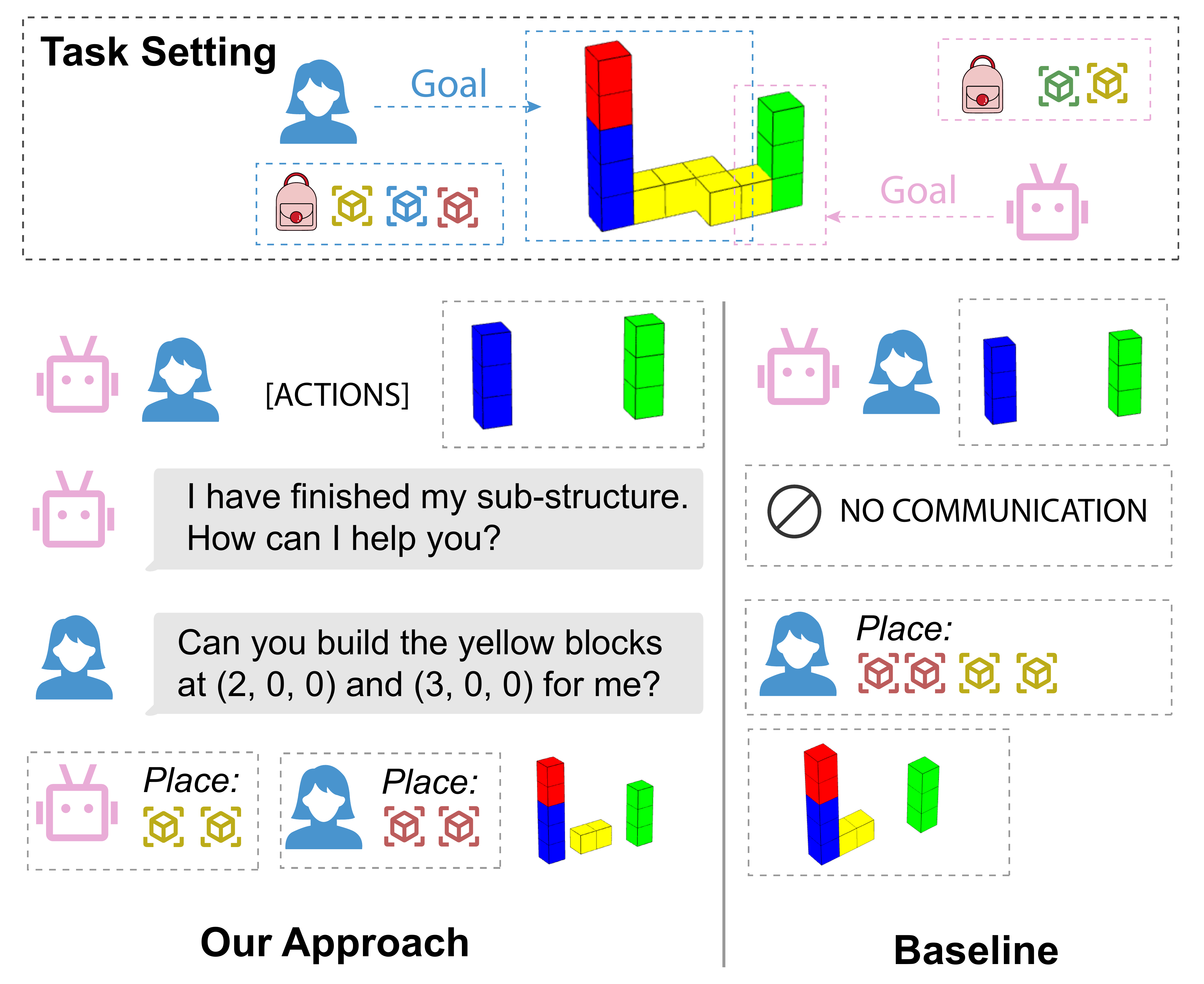} 
\caption{A human-machine example in the goal-dependent task. We show our LLM agent on the left and the baseline LLM agent on the right. Our LLM agent actively helps the partner to complete the task when the baseline keeps waiting.}
\label{fig:exp-example-2}
\end{center}
\end{figure}
\noindent \textbf{Both partner-state modeling and reflection enhance collaboration.} First, partner-state modeling significantly improves agents in the skill-dependent and goal-dependent tasks in all metrics, as LLM agents are more adept at negotiating with partners to achieve their own goals.
%
The reflection mechanism further helps adjust the communication strategy, especially when human partners are less collaborative. In 33\% skill-dependent tasks and 17\% goal-dependent tasks, human participants have a low engagement, and the LLM agents change the communication stragetgy accordingly.
For instance, one human participant continuously works on his goal even when the LLM agent asks for assistance. The LLM agent then prompts with a message that \textit{``Please build the yellow blocks from (0, 0, 2) to (0, 3, 2), so I can proceed with the task.''}. 



\noindent \textbf{LLM agents actively communicate with partners.
} 
According to Figure~\ref{fig:human-model-all} (B, C) and Figure~\ref{fig:model-only-all} (B, C), LLM agents outperform baseline agents on task workload balance and timesteps.
This is mainly due to more proactive communication. 
For example, in Figure~\ref{fig:exp-example-2}, the baseline LLM agent stops after its goal is fulfilled (observed in 56\% skill-dependent tasks and 42\% goal-dependent tasks). However, our LLM agent continues to engage, and asks, \textit{"I have finished my sub-structure. What should we do next?"}.


\noindent \textbf{Humans are smarter, but sometimes less collaborative.} 
In the human-machine experiment, a majority of humans are smarter than LLM agents—humans can actively identify and rectify LLM agents' errors (13\% skill-dependent tasks, and 8\% goal-dependent tasks).
However, a small portion of human collaborators (14.6\%) are not collaborative. In these scenarios, our reflection mechanism pushes the LLM agents to adopt a more active role in engaging with human partners.

\section{Related Work}



\noindent \textbf{Multi-Agent Collaboration with Reinforcement Learning.} Several approaches have been proposed to use reinforcement learning for multi-agent collaboration~\cite{DBLP:journals/jair/BloembergenTHK15}, including both value-based~\cite{DBLP:conf/ijcai/ChhogyalNGD19} and policy-based methods~\cite{DBLP:conf/nips/LoweWTHAM17}. These methods aim to estimate the expected value for every decision-making turn by considering both the agent itself and partner agents~\cite{DBLP:journals/jair/BloembergenTHK15}. 
Unlike existing approaches that model information exchange over a predetermined set of primitives, we focus on a more generic scenario where agents communicate in natural language.
\noindent \textbf{Agents based on LLMs.}
%
Existing research for agents based on LLMs mainly focuses on text-based, single-agent settings, such as web navigation and text games~\cite{DBLP:journals/corr/abs-2305-11854, DBLP:conf/nips/Yao0YN22, DBLP:journals/corr/abs-2307-13854, DBLP:journals/corr/abs-2303-17491, DBLP:journals/corr/abs-2309-07864}. 
Methods have been proposed to improve the planning and reasoning capacities of these agents, including chain-of-thought~\cite{DBLP:conf/nips/Wei0SBIXCLZ22, DBLP:journals/corr/abs-2303-17491,zhang2023exploring}, self-consistency decoding~\cite{DBLP:conf/iclr/0002WSLCNCZ23}, task decomposition~\cite{DBLP:conf/iclr/ZhouSHWS0SCBLC23}, and error reflection~\cite{yao2022react}.
Concurrent studies~\cite{zhou2023sotopia, gong2023mindagent} extend LLM agents into multi-agent settings but are mainly constrained to multiple LLM agents.
We build upon the recent development of LLM agents but mainly investigate the potential of multi-agent (especially human-LLM) collaboration. 
\noindent \textbf{Theory of Mind Agents.}
%
Effective collaboration requires agents to infer the intents of partner agents. Theory of Mind (ToM) refers to the agent's ability to impute mental states to the agent itself and other agents. With the help of LLMs, there are a significant amount of discussions about whether ToM has emerged by LLM agents~\cite{ma2023towards, DBLP:journals/corr/abs-2309-01660, kosinski2023theory}. We follow the idea of ToM and propose to incorporate the partner-state modeling approach into chain-of-thought prompts to build LLM agents.

\section{Conclusion}
In this paper, we introduce a new collaborative blocks environment (\envname), where human or LLM agents collaborate to complete a target structure. 
We prompt LLM agents to make decisions about the next action in \envname, and further propose to enhance collaborative abilities by modeling partner agents' state and intention and correcting errors and communication strategies from feedback. Both human-machine and machine-machine experiments indicate the effectiveness of our agents, especially on more challenging tasks that require more advanced collaboration. We believe this work provides resources and insights for future work in advancing multi-agent collaboration for social good that requires different levels of collaboration strategies. 



\newpage
\section*{Limitations}

\noindent \textbf{Multi-Agent Collaboration.} A primary limitation of our study is that we focus on a two-agent setting. 
While our framework serves as an initial study, it may not fully capture the dynamics of multi-agent collaboration~\cite{wilsker1996study}. Future work should aim to expand our platform to accommodate more than two agents and involve multiple human participants.

\noindent \textbf{Generalizability of Agent-State Modeling.} Our agent-state modeling approach mainly focuses on information specific to the blocks world. To extend our approach to other domains, we should reconfigure the agent-state format to include more diverse information about agents' states. Additionally, our design overlooks the aspect of memory, which is crucial for long-term collaboration. Incorporating a partner's memory status into decision-making can be helpful during collaboration.

\noindent \textbf{Agents Capability.} We define the agent's capabilities primarily in terms of block inventory. However, there are various other capabilities, such as breaking, placing, or picking up resources, which could significantly enrich the collaborative process. Future research should explore more diverse capabilities to provide a  better understanding of multi-agent collaboration.

\section*{Ethical Consideration}
We conduct a human-machine experiment, recruiting participants through Amazon Mechanical Turk. We get an IRB-approved protocol, ensuring no personal information from participants was recorded. We check the collected data to ensure it contains no hate speech or personal information. We implement our blocks world in the web portal with the 3D blocks world, therefore we require that no participant suffers from 3D motion sickness and would be harmed by the experiments.

There is a risk that the LLM agents might be assigned ethically questionable or unjust goals by stakeholders. 
In our current platform that is centered around block-building tasks, the possibility of assigning harmful goals is greatly minimized. However, we recognize that our proposed agents could raise ethical concerns if applied into more realistic settings like violence investigation~\cite{DBLP:conf/naacl/HuHPOKBD22}. Therefore, we emphasize the importance of designing LLM agents to be ethically responsible regarding their assigned goals. 



\bibliography{main}

\appendix

\section{Experiment Setting Details}
\label{supp:exp}
\subsection{Structure Preparation}
\label{supp:structure}

\begin{table}[ht]
\centering
\begin{tabular}{|c|c|c|c|}
\hline
\includegraphics[width=0.4\linewidth]{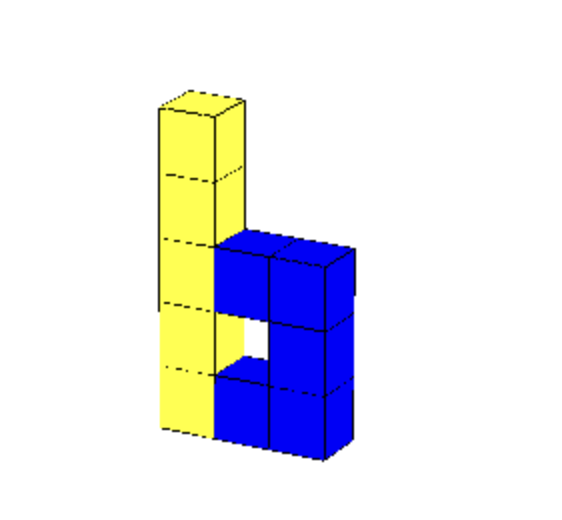} &
\includegraphics[width=0.4\linewidth]{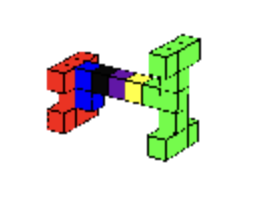} \\
Symbol & Bridge \\
\hline
\includegraphics[width=0.4\linewidth]{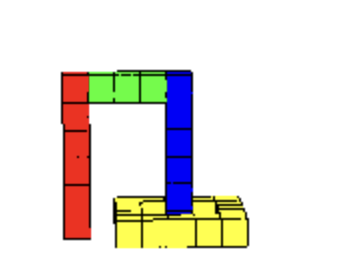} &
\includegraphics[width=0.23\linewidth]{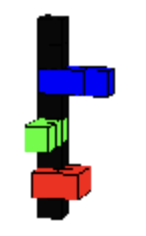} \\
Arch & Tower \\
\hline
\multicolumn{2}{|c|}{\includegraphics[width=0.45\linewidth]{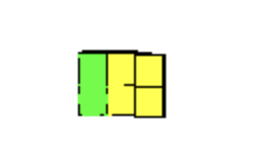}} \\
\multicolumn{2}{|c|}{Rectangle} \\
\hline
\end{tabular}
\caption{Common base structures.}

\label{tab:common}
\end{table}

We prepare the structures based on a semi-automatic pipeline.
We first identify five types of common structures: symbol, bridge, arch, tower, and rectangle (Table~\ref{tab:common}).
Then, we manually create rules for the structure types. For example, the arch’s rule is \textsc{<Two Pillars: height > 3} and width < 2; \textsc{Two pillars distance > 3 and width < 3>};
We model each surface in the target structure as the graph node and use the adjacency as the edges. Then, we represent the complexity of the structure by the number of graph-spanning trees.
To generate the structures, we randomly initialize the structure with 3 blocks. Then, we run a depth-first search following the pre-defined rules and the complexity constraint.
For the independent tasks, we use 24 structures, 16 are manually created and 8 are created based on the pipeline. The skill-dependent structures share the same structures but with different inventories; For goal-dependent tasks, we use 24 structures, 16 are manually created and 8 are created based on the pipeline.


\subsection{Steps in User Manual}
\label{supp:manual}

\begin{table}[ht]
\centering
\begin{tabular}{|c|c|c|c|}
\hline
\includegraphics[width=0.4\linewidth]{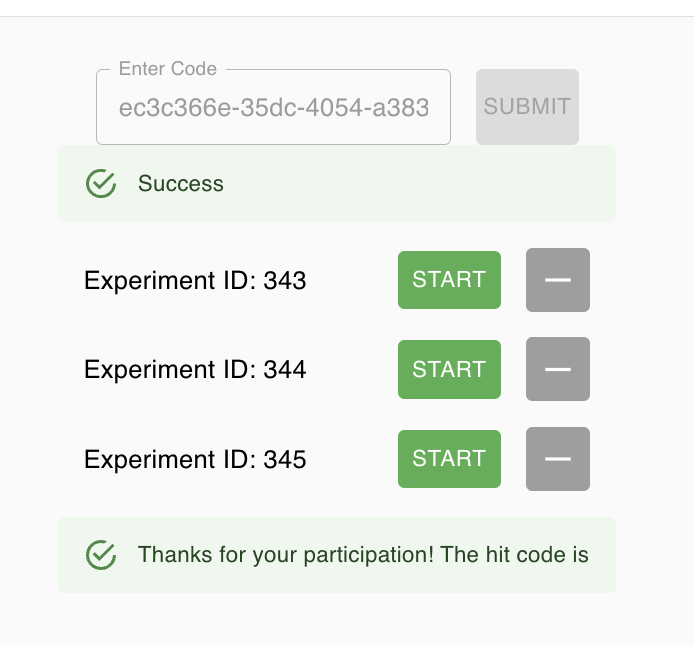} &
\includegraphics[width=0.4\linewidth]{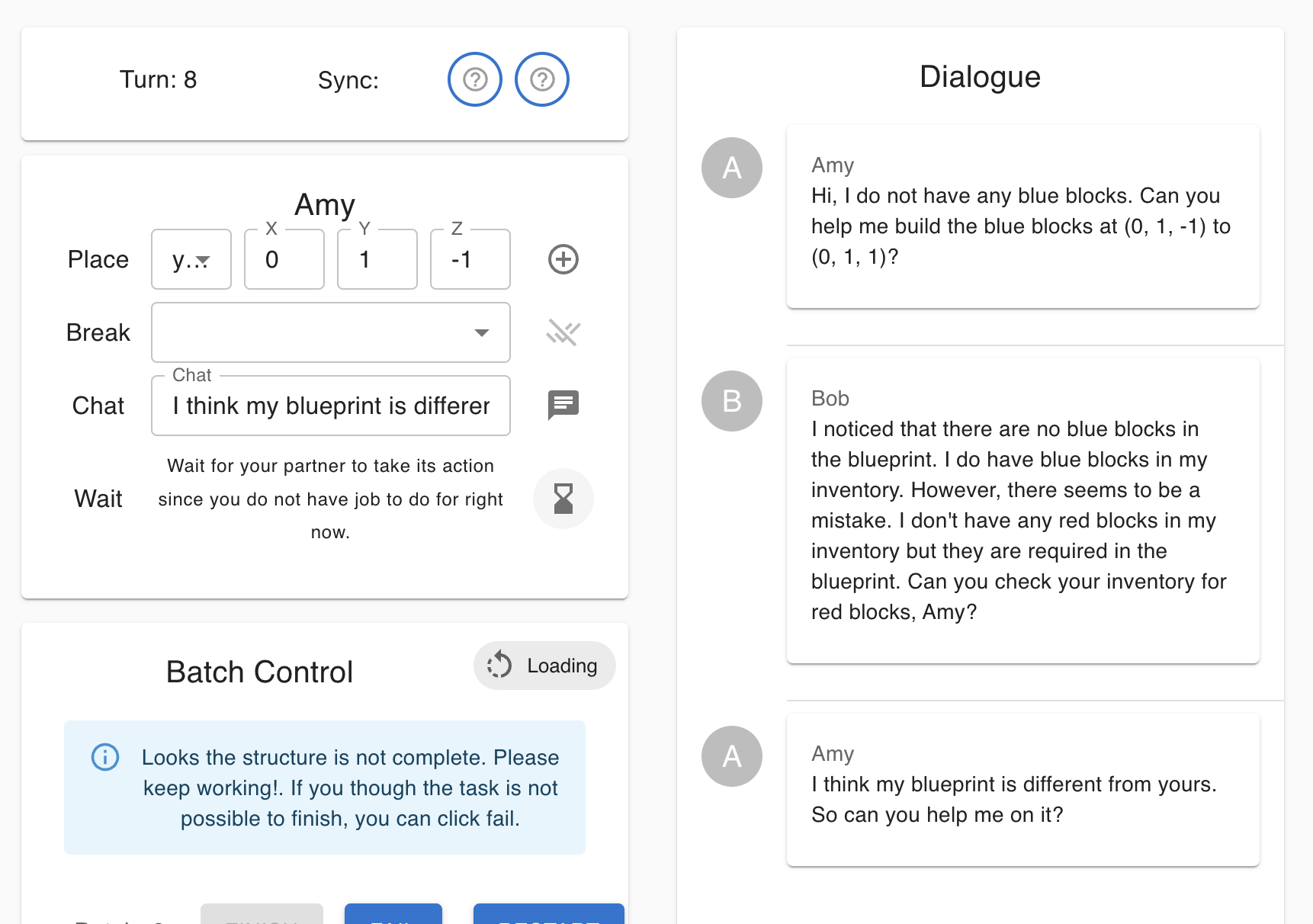} \\
Step 1 & Step 2 \\
\hline
\includegraphics[width=0.4\linewidth]{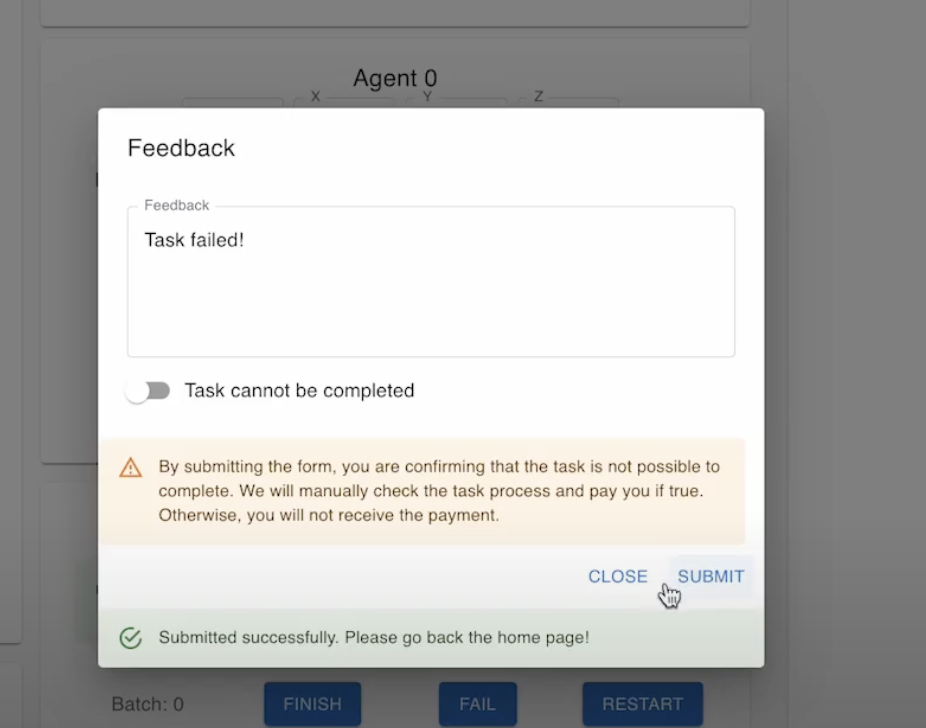} &
\includegraphics[width=0.23\linewidth]{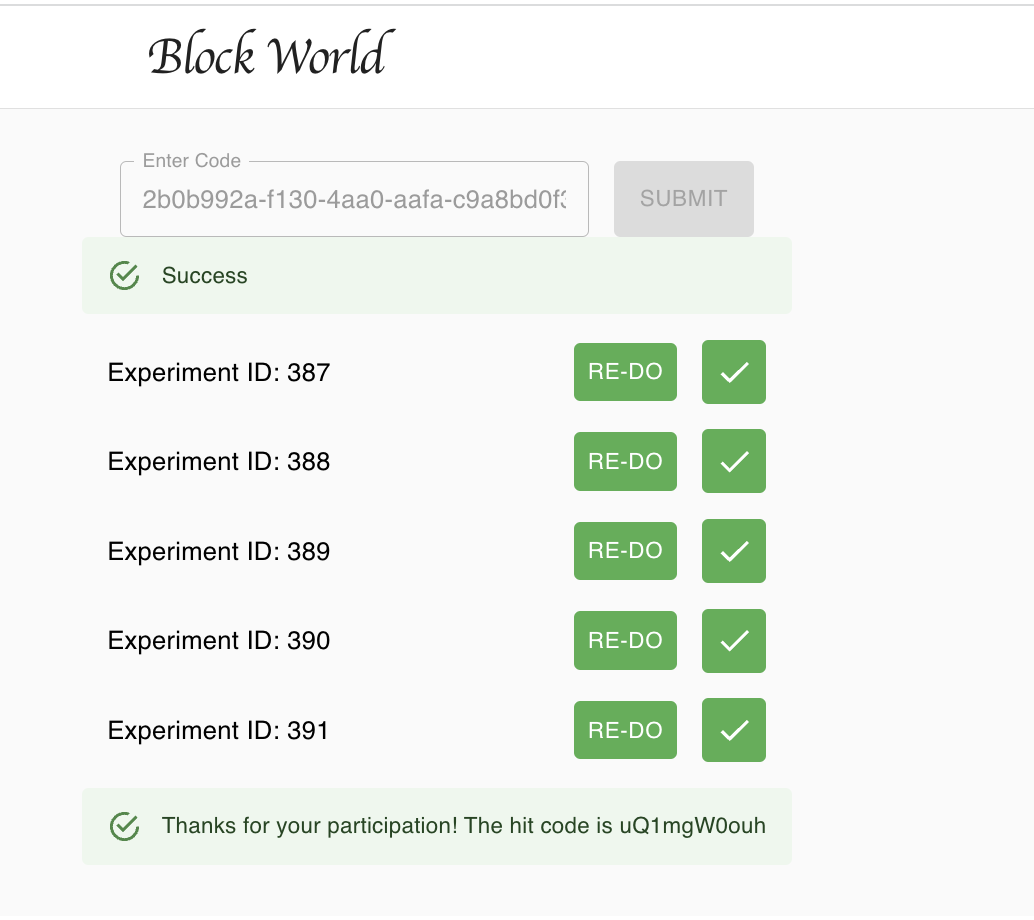} \\
Step 3 & Step 4 \\
\hline
\end{tabular}
\caption{Selected steps in the user manual.}

\label{tab:manual}
\end{table}

The following steps are outlined in the user manual. We also attach the figures of the steps in table~\ref{tab:manual}.
\begin{enumerate}
\item Users are assigned a unique participant ID and a general web URL for access. Upon opening the website, they should enter their assigned participant code to begin the experiments. The webpage will display the assigned tasks.
\item Users can then click on each task displayed on the page to start. In these tasks, users are required to collaborate with LLM agents to complete structure-building activities.
\item Upon task completion, \envname will automatically notify the user, who then closes the page to proceed to the next task. If the user claims a task is impossible to complete (possibly due to issues with the LLM agent), they can click the submit button and provide a reason for the task’s incompletion.
\item After finishing a task, the rightmost icon on the page will turn green, indicating completion. Once all tasks are completed, a success code will appear. Users should copy this code and submit it to Amazon Mechanical Turk for payment.
\end{enumerate}

\subsection{Institutional Review Board}
Our study is conducted under the approval of our university's Institutional Review Board (IRB). We did not collect any identity or demographic information about the crowd-sourcing workers. We will disclose the IRB information after the paper is accepted.
\subsection{Workload Balance Evaluation}
To evaluate the workload balance, we need first to compute the optimal assignment of the workload balance. 
Shown in Algorithm~\ref{alg:optimal_wb}, we first identify the mutual blocks ($e^m$) shared between the two agents' inventories ($e^1$ and $e^2$). Unique blocks for each agent are then determined by subtracting these mutual blocks from their respective inventories, resulting in $e^1_{unique}$ and $e^2_{unique}$. These unique blocks are directly assigned to the corresponding agent. The core of the algorithm focuses on the equitable distribution of mutual blocks. Blocks are assigned from $e^m$ to the agent with the lesser total block count, ensuring a balanced workload. This process continues until all mutual blocks are evenly distributed or depleted. The algorithm thus optimizes task allocation by balancing the number of blocks each agent is responsible for, in alignment to achieve an optimal workload balance ($\gamma$) between the agents in constructing the target structure ($G$).

\begin{algorithm}[H]
\caption{Compute Optimal Workload Assignment}
\begin{algorithmic}[1]
\State \textbf{Input:} Agent inventories $e^1$, $e^2$; Target structure $G$
\State \textbf{Output:} Optimal assignment of blocks to agents

\Procedure{OptimalAssignment}{$e^1$, $e^2$, $G$}
    \State $e^m \gets e^1 \cap e^2$ \Comment{Identify mutual blocks}
    \State $e^1_{unique} \gets e^1 \setminus e^m$ \Comment{Unique blocks for Agent 1}
    \State $e^2_{unique} \gets e^2 \setminus e^m$ \Comment{Unique blocks for Agent 2}
    \State Assign $e^1_{unique}$ and $e^2_{unique}$ to corresponding agents
    
    \While{$e^m$ is not empty}
        \State Calculate $count^1$ and $count^2$ \Comment{Current block count for each agent}
        \If{$count^1 \leq count^2$}
            \State Assign block from $e^m$ to Agent 1
            \State Remove the assigned block from $e^m$
        \ElsIf{$count^2 < count^1$}
            \State Assign block from $e^m$ to Agent 2
            \State Remove the assigned block from $e^m$
        \EndIf
    \EndWhile
    \State \textbf{return} Assignment of $e^1_{unique}$, $e^2_{unique}$, and balanced $e^m$ to agents
\EndProcedure
\end{algorithmic}
\label{alg:optimal_wb}
\end{algorithm}

\section{Prompt Text}
\label{sec:prompt}
In this appendix, we present the prompt used in our study. It is important to note that the concepts included in our actual prompt vary from those discussed in the main paper. To clarify these differences, we have outlined the relationships between the concepts used in the actual prompt and those in the paper in Table~\ref{tab:concepts}.
\begin{table*}[ht]
\begin{center}
    
\begin{tabular}{|l|l|p{7cm}|}
\hline
Prompt                   & Paper                 & Explanation \\ \hline
Minecraft               & CoBlock World   & Initially, CoBlock was designed to mimic the Minecraft-styled blocks world, therefore, we use Minecraft to help LLM agents to understand our environment.             \\ \hline
Motive                  & Goal               &   The motive is identical to the goal used in the paper.          \\ \hline
Visual Motive           & N/A                    &     During the early phase of the design, we aim to support the visual and textual description of the goals.  Therefore, we also prompt LLM agents to understand the textual goals. However, in the experiments, we did not use the textual description, which will be explored in the future.      \\ \hline
Textual Motive          & N/A                    &             \\ \hline
Theory-of-mind Modeling & Partner-State Modeling &    The theory-of-mind modeling refers to partner-state modeling and the understanding of the self status. In the paper, we highlight the partner-state modeling because the self-status modeling obviously exists in the first step of the prompt and the baseline prompt.       \\ \hline
\end{tabular}
\end{center}
\caption{Concept comparison in the prompt text and paper.}
\label{tab:concepts}
\end{table*}

\subsection{Task Description Prompt}
The task description is presented to all LLM agents of both the baseline prompt and our approach. It details the environment input/output formats to ensure the LLM agents understand our \envname environment. The task description consists of the task summary, world state, inventory, message, and goal formats.
\createpromptbox{Task Summary}{
\prompttext{
I want you to act as a Minecraft player collaborating with another agent to build a structure with a blueprint. You need to use the following commands to interact with the Minecraft world:
}
\prompttext{\# Place a red block at the position of (0, 1, 1). }
\textit{place$\_$block(block$\_$type=red, pos=(0,1,1)) }\\
\prompttext{\# chat with your partner}
\textit{send$\_$message(message=``Hello, partner'')}\\
\prompttext{\# destroy the block at the position of (3, 1, 3)}
\textit{break$\_$block(pos=(3, 1, 3))}
}

\createpromptbox{World State Format}{\prompttext{
At each turn, you will receive the following information:
}
\prompttext{\# World state: You will get the position of all blocks in the world in the following format:}
\prompttext{\# Please note that the ground is the y=1 plane.}
\prompttext{\textit{<World>}}
    \inpromptblock{red}{(0, 1, 2)}\\
    \inpromptblock{yellow}{(0, 1, 3)}\\
    \inpromptblock{purple}{(0, 1, 4)}\\
\textit{<\/World>}
}

\createpromptbox{Inventory Format}{
\prompttext{\# You will get your inventory in the following format: (each time you place a block from the inventory in the world, you will lose it in the inventory)}
\prompttext{\textit{<Inventory>}\\
    \inpromptinventory{red}{3}\\
    \inpromptinventory{yellow}{3}\\
\textit{<Inventory>}}
}

\createpromptbox{Message Format}{
\prompttext{\# You will get the message history between you (ChatGPT) and your partner in the following format:}
\prompttext{\textit{<Dialogue>}\\
    \textit{<sender=``ChatGPT'', message=``Hello!''>}\\
    \textit{<sender=``Partner'', message=``Hi, I am your partner!''>}\\
\textit{<Dialogue>}}
}

The goal is represented through a list of blocks that need to be constructed. In our system development, we aim to represent goals using both textual and visual modalities. However, this paper focuses on the visual representation of goals. The goal is articulated as a motive in our prompts. The visual motive is delineated by an optional description of the target structure and a comprehensive list of the blocks constituting that structure. It is important to note that although the structure's shape must adhere to specified guidelines, its location can be decided by the agent.

\createpromptbox{Goal (Motive) Format}{
\prompttext{<Motives>}
\prompttext{\# You will get a list of motives you need to fulfill. There can be two types of motives, i.e. visual and textual. }
\prompttext{<Motives>}
%
\prompttext{<VisualMotive>\\
    <Description> A simple two-layer structure consisting of red and yellow blocks \\</Description>}
    \inpromptblock{red}{(0, 1, 2)}\\
    \inpromptblock{red}{(1, 1, 2)}\\
    \inpromptblock{red}{(0, 1, 3)}\\
    \inpromptblock{red}{(1, 1, 3)}\\
    \inpromptblock{yellow}{(0, 2, 2)}\\
    \inpromptblock{yellow}{(1, 2, 2)}\\
    \inpromptblock{red}{(0, 2, 3)}\\
    \inpromptblock{yellow}{(1, 2, 3)}

\prompttext{</VisualMotive>\\
}
}

\subsection{In-Context Examples }
Below, we present the in-context examples used in our prompt. Since \envname~is a multi-round game, our examples span three rounds, enabling LLM agents to plan across multiple rounds.
Please notice that the prompt will be provided after the task description when querying LLMs.
\createpromptbox{Round 1 Input}{

\prompttext{<Input>}
\prompttext{<Motive><TextualMotive 
text="Construct a bridge with a span of 12 blocks using green blocks. Place two green block pillars, each consisting of 4 blocks. Add 4 yellow blocks as the bridge surface. ”
/>
</Motive>}
\prompttext{
<World>\\
</World>\\
<Inventory>\\
     <block $block\_type $=``yellow'', count=20>\\
     <block $block\_type $=``green'', count=20>\\
     <block $block\_type $=``purple'', count=20>\\
</Inventory>\\
<Dialogue>\\
</Dialogue>
}
\prompttext{</Input>}
}

\createpromptbox{Round 1 Expected Output}{\prompttext{\# Partner Modelling}
\prompttext{\# Long-term goal: Unknown}
\prompttext{\# Short-term goal: Unknown}
\prompttext{\# Partner inventory: Unknown}
\prompttext{\# Explanation: I do not know anything about my partner.}

\prompttext{\# Self Modelling}
\prompttext{\# Long-term goal: creating the target structure}
\prompttext{\# Short-term goal: Build the green pillars}
\prompttext{\# My inventory: [yellow: 20, green: 20, purple: 20]}
\prompttext{\# Explanation: Since my long-term goal is to build the bridge, I need to decompose it into different small parts. The first part I can build is the pillars.}
\prompttext{\# Since I do not have any access to the partner’s information, I need to let him know my intention and inventory when asking for it. }
\prompttext{send$\_$message(message=``Hi, I need to build a bridge with the yellow deck and green pillars. I have green, yellow and purple. How about you?'')}
}

\createpromptbox{
Round 2 Input
}{
\prompttext{<Input>}
\prompttext{<Motives>}
\prompttext{<TextualMotive 
text=``Construct a bridge with a span of 8 blocks and a width of 3 blocks using yellow blocks. Place two green block pillars, each consisting of 4 blocks. Add 4 yellow blocks as the bridge surface.''
/>}\\
</Motives>\\
<World>\\
\inpromptblock{red}{(0, 1, 2)}\\
\inpromptblock{red}{(1, 1, 2)}\\
\inpromptblock{red}{(0, 1, 3)}\\
\inpromptblock{red}{(1, 1, 3)}
\prompttext{</World>}
\prompttext{<Inventory>}
    \inpromptinventory{yellow}{20}\\
    \inpromptinventory{green}{20}\\
    \inpromptinventory{purple}{20}\\
\prompttext{</Inventory>}
\prompttext{<Dialogue>
    <chat sender=``Agent 1'', message=``Hi, I need to build a bridge with the yellow deck and green pillars. I have green, yellow and purple. How about you?''/>
    <chat sender=``Agent 2'', message=``Hi, I have red, green, and black. I need to build the black fence upon your deck.''/>
</Dialogue>}
\prompttext{</Input>}}

\createpromptbox{Round 2 Expected Output}{
\prompttext{\# Partner Modelling}
\prompttext{\# Long-term goal: Build the fence on the deck}
\prompttext{\# Short-term goal: Wait until the fence is built}
\prompttext{\# Partner inventory: [red: unknown, green: unknown, black: unknown]}
\prompttext{\# Explanation: I have extracted the partner information from the partner’s response.}

\prompttext{\# Self Modelling}
\prompttext{\# Long-term goal: creating the target structure}
\prompttext{\# Short-term goal: Build the green pillars. Ask for the partner’s help with the pillars}
\prompttext{\# My inventory: [green: 20, red: 20, purple: 20]}
\prompttext{\# Explanation: Since my long-term goal is to build the bridge, I need to decompose it into different small parts. The first part I can build is the pillars. I can ask for the partner’s help because it has the green block in the inventory.}

\prompttext{$send\_message$(message="Hi, I need to build the two pillars first. My bridge spans 12 and I need to have 2 pillars. So it would be reasonable to put the two pillars at (4, 0, 0) and (8, 0, 0). I will start with the (4, 0, 0) one and could you please build the pillar at (8, 0, 0) ?")}

\prompttext{\# Since the motive does not specify the locations of the pillars, I need to first decide them. Given the fact that the partner is waiting and can help me build the pillar, it is also reasonable to ask for its help.}
\prompttext{\# Build the yellow pillar on (4, 0, 0).}

\prompttext{$place\_block$($block\_type $="yellow", pos=(4, 0, 0))}
}

\createpromptbox{Round 3 Input}{
\prompttext{<Input>}
\prompttext{<Motives>}
\prompttext{<TextualMotive 
text="Construct a bridge with a span of 8 blocks and a width of 3 blocks using yellow blocks. Place two green block pillars, each consisting of 4 blocks. Add 4 yellow blocks as the bridge surface. ”
/>}
\prompttext{</Motives>}
\prompttext{<World>}
    \inpromptblock{yellow}{(4, 0, 0)}\\
    \inpromptblock{yellow}{(4, 1, 0)}\\
    \inpromptblock{yellow}{(4, 2, 0)}\\
    \inpromptblock{yellow}{(8, 1, 0)}\\
    \inpromptblock{yellow}{(8, 2, 0)}\\
    \inpromptblock{yellow}{(8, 3, 0)}\\
    
\prompttext{</World>}
\prompttext{<Inventory>}
\inpromptinventory{yellow}{20}
\inpromptinventory{green}{20}
\inpromptinventory{purple}{20}

\prompttext{</Inventory>}
\prompttext{<Dialogue>}
\prompttext{<chat sender=``Agent 1", message="Hi, I need to build a bridge with the yellow deck and green pillars. I have green, yellow and purple. How about you?''/>}
\prompttext{<chat sender=``Agent 2", message="Hi, I have red, green and black. I need to build a black fence upon your deck''/>}
\prompttext{<chat sender=``Agent 3", message="Sure, I will build the yellow pillar as you requested.''/>}
\prompttext{</Dialogue>}
\prompttext{</Input>}
}

\createpromptbox{Round 3 Expected Output}{\prompttext{\# Partner Modelling}
\prompttext{\# Long-term goal: Build the fence on the deck}
\prompttext{\# Short-term goal: Building the pillar at (8, 0, 0) as I requested.}
\prompttext{\# Partner inventory: [red: unknown, green: unknown, black: unknown]}
\prompttext{\# Explanation: I have extracted the partner's information from the partner’s response.}

\prompttext{\# Self Modelling}
\prompttext{\# Long-term goal: creating the target structure}
\prompttext{\# Short-term goal: Build the yellow pillar on (4, 0, 0)}
\prompttext{\# Explanation: Since the yellow pillar has not been finished, I need to continue building it.}

\prompttext{$place\_block$($block\_type $=``yellow'', pos=(4, 4, 0))}
\prompttext{\# Build the last block of the pillar as it is my short-term goal.}

\prompttext{$place\_block$($block\_type $=``yellow'', pos=(4, 0, 0))}
\prompttext{$send\_message$(message``I have completed my pillar. Then I will start building the green deck.!'')}
\prompttext{$place\_block$($block\_type $=``green'', pos=(4, 5, 0))}
\prompttext{$place\_block$($block\_type $=``green'', pos=(3, 5, 0))}

\prompttext{\# Since I have finished the pillar and can assume my partner will finish his assigned pillar. I can start building the green deck.}
}

\end{document}